\documentclass[runningheads]{llncs}
\usepackage[T1]{fontenc}
\usepackage{graphicx}
\usepackage{amsmath} 
\usepackage{amssymb}
\usepackage{booktabs}
\usepackage{siunitx}
\usepackage{makecell}
\usepackage{hyperref}   
\usepackage{cleveref}   
\crefname{figure}{Fig.}{Figs.}
\crefname{table}{Tab.}{Tabs.}
\crefname{equation}{Eq.}{Eqs.}
\crefname{section}{Sect.}{Sects.}
\usepackage{color}

\begin{document}
\title{CLANE: Continual Learning of Actions on Neuromorphic Hardware from Event Cameras}
\titlerunning{CLANE}
%
\author{
Elvin Hajizada\inst{1\textsuperscript{†}}\and
Michael Neumeier\inst{2,3\textsuperscript{†}}\and
Edward Paxon Frady\inst{4} \and 
Yulia Sandamirskaya\inst{5}\and
Axel von Arnim\inst{2}\and
Bing Li\inst{6} \and
Eyke Hüllermeier\inst{1,7,8}}


\authorrunning{E. Hajizada et al.}

%
\institute{Institute of Informatics, University of Munich (LMU), Munich, Germany\and fortiss GmbH, Neuromorphic Computing, Munich, Germany \and Technical University of Munich, TUM School of CIT, Munich, Germany \and Intel Labs, Intel Corporation, Santa Clara, CA, USA \and Institute of Computational Life Sciences (ICLS), Zurich University of Applied Sciences (ZHAW), Wädenswil, Switzerland \and Technische Universität Ilmenau, Resource-Efficient Artificial Intelligence Group, Ilmenau, Germany \and Munich Center for Machine Learning (MCML) \and German Research Centre for Artificial Intelligence (DFKI), Kaiserslautern, Germany \\
\email{hajizada.elvin@campus.lmu.de}\\
$^\dagger$Equal contribution.}
\maketitle              

\begin{abstract}
Recognizing and continuously learning novel human actions without forgetting prior classes is a requirement for emerging AR/VR and robotics applications. For these applications, both on-device processing and learning are essential for privacy and low-latency adaptation. Event cameras address the efficiency of visual sensing  with sparse, asynchronous output that is naturally compatible with neuromorphic processing. Yet no prior system has deployed a continual on-device learning pipeline for event-based action recognition using neuromorphic hardware. We present CLANE -- \underline{C}ontinual \underline{L}earning of \underline{A}ctions on \underline{N}euromorphic hardware from \underline{E}vent cameras -- deployed end-to-end on Intel Loihi 2. CLANE combines a spiking 2D CNN for spatiotemporal feature extraction with CLP-SNN as its on-chip learning head, extended to action clips via a Temporal Aggregation Layer and a fixed-point Normalization Layer -- both novel Loihi 2 modules. On THU\textsuperscript{E-ACT}-50 dataset (50 classes, real-world conditions), CLANE achieves 70.4\% accuracy at a continual learning task while delivering >100× energy reduction and 16× lower latency over a sequential CNN+GRU+CLP edge GPU baseline, validated through iso-algorithm cross-platform benchmarking across three evaluation levels.

\keywords{Neuromorphic Computing  \and Continual Learning \and Action Recognition \and Event-based Cameras \and Cross-hardware Benchmarking.}
\end{abstract}
\section{Introduction}
\label{sec:intro}

Continual learning from experience -- acquiring new knowledge sequentially without forgetting what came before -- is a routine capability of the brain, yet engineering it into always-on, power-constrained edge devices remains a significant open challenge~\cite{de2021CLreview}.
A key reason is architectural: standard deep learning requires batch retraining via global error propagation, large activation buffers, and access to past data or large replay buffers, all incompatible with the power and memory budgets of edge devices~\cite{french1999catastrophic}.
Biological nervous systems, on the other hand, update synaptic weights through local three-factor plasticity rules, depending only on pre- and post-synaptic activity and diffuse modulatory signals available at the synapse~\cite{fremaux2016neuromodulated}.
Global error propagation, by contrast, requires transmitting loss gradients across all layers and storing intermediate activations -- operations with no plausible biological substrate~\cite{lillicrap2020backpropagation} that stretch both compute and memory of on-device computers. 
Neuromorphic processors draw inspiration from the brain and implement event-driven spike routing, near-memory computing, and sparse parallel computation~\cite{schuman2022neuromorphicSurvey}.
Intel Loihi 2 neuromorphic chip also provides on-chip learning acceleration, computing weight updates close to where the weights reside, avoiding DRAM transfers that dominate GPU energy budgets~\cite{davies2018loihiLearning,orchard2021loihi}.
This makes neuromorphic hardware a natural candidate for energy-efficient continual learning -- provided the learning algorithm itself respects locality constraints.

Real-world deployment scenarios for such a system include AR/VR and robotics, where action or gesture recognition must operate in real time at the edge under strict power budgets. The challenges of visual action perception, such as variable lighting and temporal granularity, leading to redundant data in static scenes and motion blur during fast movement, can be addressed by event-based cameras -- retina-inspired sensors that asynchronously report per-pixel brightness changes as sparse spike trains with microsecond precision and high dynamic range~\cite{gallego2020eventSurvey}.  These sensors communicate only pixels with intensity changes, producing representations that are inherently sparse and architecturally compatible with the event-driven computation of neuromorphic processors, enabling low-latency, low-energy action recognition systems.

A further challenge of on-device action recognition is that the space of possible human actions is large, unstructured, and evolving~\cite{shen2024towards}. Users introduce novel gestures, environmental conditions shift, and recognition systems must adapt continuously without forgetting previously learned actions. This demands online, class-incremental CL directly on the deployed device, rather than periodic retraining from shuffled batches~\cite{hayes2022online}. The on-chip local learning acceleration properties of Loihi 2 make it a natural execution substrate for exactly this regime.

\begin{figure}[t]
\centerline{\includegraphics[trim={0 2.2cm 1.0cm 0.7cm},clip,width=\textwidth]{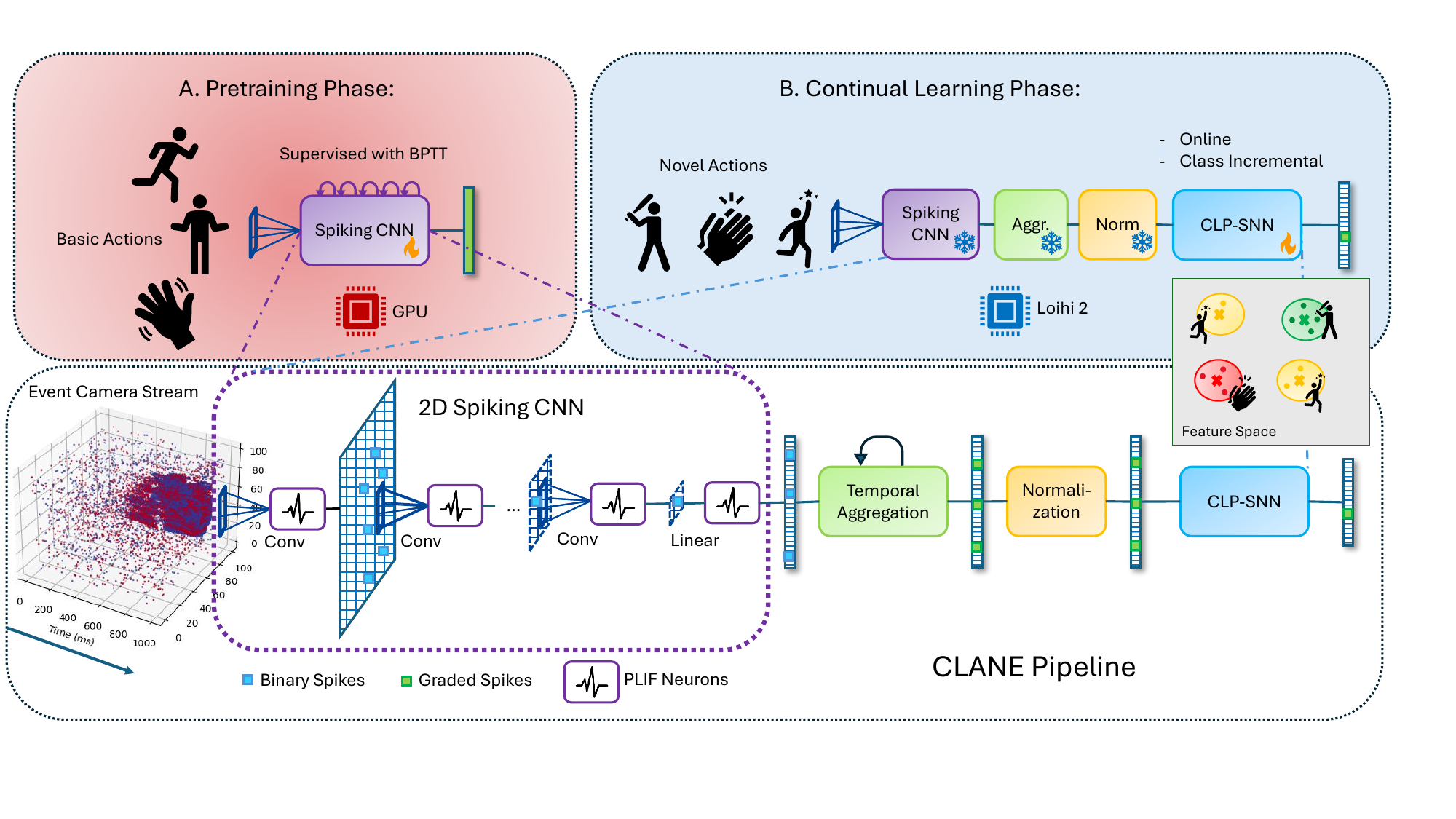}}
\caption{Our CLANE pipeline with Spiking CNN, Temporal Aggregation, Normalization, and CLP-SNN including sparse binary or graded spike activity maps and vectors. Schematic visualization of pretraining in A and continual learning in B.}
\vspace{-5pt}
\label{fig:architecture}
\end{figure}

Spiking neural networks (SNNs) complete the brain-inspired stack outlined above. Their stateful neurons temporally integrate incoming spike streams and communicate via asynchronous spikes~\cite{yamazaki2022snn}, making them a natural computational fit for the sparse, asynchronous data produced by event cameras~\cite{neumeier2025,vicente2025}.
Together, event cameras, SNNs, and neuromorphic hardware form a stack where sparsity generated at the sensor propagates through computation—a three-way architectural alignment that serves as the co-design principle~\cite{schuman2022neuromorphicSurvey}, central to our work.
This alignment extends naturally to learning that is event-driven, sparse, and local.
Yet despite on-chip learning mechanisms being available on chips like Loihi 2, deploying this full stack with a principled continual learning (CL) algorithm for large-scale spatiotemporal tasks remains unrealized.

Previous work has taken steps toward this goal by demonstrating few-shot gesture learning on Loihi 1, though limited to a small benchmark and without the CL and forgetting analysis~\cite{stewart2020online}. More recently, Hajizada et al. proposed CLP-SNN, a neuromorphic CL method implemented on Loihi 2 for object recognition~\cite{hajizada2025}. This work was limited to static spatial inputs. More broadly, no prior neuromorphic CL work has been evaluated on a large-scale, real-world dataset approaching the complexity of full-body action recognition. In this regard, extending CLP-SNN to large-scale spatiotemporal recognition tasks poses three unresolved challenges: (1) CLP-SNN processes fixed-size static feature vectors and cannot directly handle temporal spike sequences produced by a deep SNN over an action clip duration; (2) CLP-SNN requires L2-normalized input vectors, yet Loihi 2 provides no native division instruction, making on-chip normalization a non-trivial hardware design problem; (3) the neuromorphic CL literature lacks a benchmarking methodology that simultaneously controls for model capacity, learning algorithm, and hardware -- without which efficiency claims cannot be cleanly attributed to hardware and algorithmic differences.

We present CLANE (Continual Learning of Actions on Neuromorphic Hardware from Event Cameras), an end-to-end SNN with CL capability deployed on Intel Loihi 2 addressing all the above (\cref{fig:architecture}).
CLANE integrates a 2D spiking CNN for spatiotemporal feature extraction with CLP-SNN~\cite{hajizada2025} as its on-chip continual learning head, learning novel action classes online and class-incrementally from THU\textsuperscript{E-ACT}-50 dataset~\cite{gao2023} in real time.
Our contributions are:
\vspace{-18pt}
\begin{itemize}
    \item \textbf{CLP-Loihi}: Extension of CLP-SNN to spatiotemporal action recognition with Temporal Aggregation and Normalization Layers, implemented on Loihi 2 via custom neuron models.
    \item \textbf{Full neuromorphic deployment}: Co-deployment of a deep spiking CNN feature extractor and the CLP-Loihi on a single Loihi 2 chip, directly processing event-camera-derived spike inputs without a GPU in either the inference or learning loop, evaluated on THU E-ACT-50 under real-world conditions including lighting variation, viewpoint changes, and occlusions.
    \item \textbf{Rigorous iso-algorithm cross-platform benchmarking} against Nvidia Jetson Orin Nano across three levels: full pipeline, learning algorithm, and feature extractor. This addresses challenge (3) by isolating the hardware contribution from algorithmic and model-capacity confounds. CLANE achieves more than $100\times$ energy reduction and $16\times$ latency improvement over a 2D CNN+GRU baseline with the same CLP algorithm on the GPU, at an accuracy cost of only 2.6\%.
\end{itemize}

\section{Related Work}
\label{sec:related_work}
\subsection{Event-based Action Recognition}
Event-based cameras have been widely applied to vision tasks such as optical flow estimation, tracking, and SLAM, where high temporal resolution and sparse events offer advantages over frame-based cameras~\cite{gallego2020eventSurvey}. 
However, most research on event-based action recognition converts event streams into dense event frames and processes these with 2D CNNs plus recurrent layers~\cite{innocenti2020}, 3D CNNs~\cite{gao2023}, or Transformers~\cite{blegiers2023}. 
These dense-tensor methods cannot exploit event sparsity and are incompatible with the event-driven computation of neuromorphic hardware.

Alternatively, SNNs can achieve temporal feature extraction without recurrent connections by leveraging voltage integration dynamics~\cite{vicente2025}. 
They have shown competitive performance, even surpassing deeper 3D CNNs~\cite{neumeier2025}. As directly mappable to neuromorphic processors~\cite{orchard2021loihi}, they yield measurable efficiency gains over GPU-based methods for sparse, event-driven workloads~\cite{neumeier2025,hajizada2025}.

Most of these studies focused on recognizing a predefined set of actions by training in a supervised, offline fashion~\cite{neumeier2025,vicente2025}. 
Stewart et al.~\cite{stewart2020online} demonstrated few-shot gesture learning using an SNN with output-layer learning capabilities and evaluated it on the first generation of Intel Loihi. Similar to our approach, this work also utilized a frozen feature extractor and a plastic learning layer. However, it was evaluated on a simpler dataset (DVSGestures~\cite{amir2017dvsGestures}) and did not include an analysis of catastrophic forgetting, which is critical for continual learning scenarios. This leaves a gap: no prior neuromorphic system has evaluated class-incremental action recognition with catastrophic forgetting analysis at a dataset scale comparable to THU\textsuperscript{E-ACT}-50~\cite{gao2023}.

\subsection{Neuromorphic Continual Learning}
Neuromorphic continual learning (NCL) has emerged as a paradigm combining event-driven neuromorphic hardware and SNNs with local plasticity mechanisms to enable real-time sequential learning without catastrophic forgetting~\cite{minhas2025continual}. This paradigm is motivated by the observation that neuromorphic hardware can support on-chip, local, and sparse weight updates, making it a natural candidate for low-power and low-latency embedded continual learning in dynamic environments~\cite{minhas2025continual}. 
Early NCL approaches augmented unsupervised STDP with adaptive learning rates, weight decay, and threshold adaptation to achieve selective learning and homeostatic stabilization~\cite{panda2017asp}.
Pes et al.~\cite{pes2024activedend} added task-specific sub-networks and context-dependent dendritic modulation. Han et al.~\cite{han2023DSD-SNN} instead proposed to dynamically grow new neurons for novel tasks and prune redundant neurons to prevent catastrophic forgetting. Finally, memory replay has also been applied to SNNs to mitigate forgetting by storing and interleaving exemplars from previous tasks~\cite{proietti2023snnreplay,minhas2025replay4ncl}.
These previous evaluations addressed spatial tasks, while still relying mostly on toy benchmarks, shallow networks, leaving real-time action learning from event cameras largely unexplored. 

Furthermore, most NCL evaluations are conducted in software simulation rather than on physical neuromorphic hardware~\cite{minhas2025continual}. System-level design of CL algorithms tailored for specific neuromorphic platforms for measurable gains in energy, latency, and memory remains underdeveloped. Recent work~\cite{hajizada2025} demonstrated NCL feasibility on Intel Loihi 2 for object recognition, by introducing CLP-SNN, a spiking extension of the Continually Learning Prototypes (CLP) algorithm~\cite{hajizada2024}. 
CLP-SNN utilizes local prototype-based representations, a self-normalizing three-factor learning rule, metaplasticity for consolidation, and on-demand neurogenesis, enabling real-time continual learning. 
It exceeds replay-based baselines in accuracy while improving the Pareto frontier by reducing latency and energy consumption~\cite{hajizada2025}. However, this method has been limited to spatial recognition tasks. The adaptation of CLP-SNN to spatiotemporal tasks and its integration with a deep convolutional SNN for event-based action recognition remains unaddressed.

\section{Methods}
\label{sec:method}

\subsection{Event-Driven Data Pipeline}
THU\textsuperscript{E-ACT}-50 dataset used the CeleX-V event camera ($1280\times800$ resolution)\cite {gao2023}, that produces asynchronous per-pixel brightness changes encoded as (x, y, t, p) tuples, where x,y indicate the location on the pixel array, t timing of the event, and p the polarity ($\pm 1$).
We select the central $600\times600$ region and bin the streams into fixed temporal windows (40ms, 10ms, or 2ms) depending on target frequency (25Hz, 100Hz, or 500Hz) and at a spatial resolution of $100\times100$, creating sparse 2D event-count histograms (event frames) per polarity, with 80\% sparsity even at 25 Hz.
This representation is compatible with Loihi 2, as only non-zero pixel values are encoded as spikes, enabling event-driven computation, whereas GPUs' dense processing of all pixels, regardless of sparsity, leads to inefficiency with sparse event data~\cite{orchard2021loihi}. 

\subsection{Spiking Convolutional Feature Extractor}
\textbf{Architecture:} The resulting stream of sparse event frames is fed into a spiking convolutional neural network for feature extraction.
The neural network consists of 5 spiking convolutional layers, followed by a flattening operation and one spiking fully connected layer (\cref{fig:architecture}). Incoming spikes $s(t)$ are processed by a 2D convolutional kernel ($Conv_{3\times3}$) and batch normalization ($BN$):
\[
x(t) = BN(Conv_{3\times3}(s(t))) 
\]
The resulting neuronal input current $x(t)$ is accumulated together with a bias $b(t)$ to the neuron's membrane potential $v(t)$ following Parametric LIF~\cite{fang2021} dynamics, spiking $s(t)$ and reset functions, with $\alpha_v$ as trainable decay parameter per layer:
\begin{align*}
  v(t) &= \alpha_v v(t-1) + x(t) + b(t), &
  s(t) &= (v(t) \geq \vartheta), &
  v(t) &= v(t)\,(1-s(t))
\end{align*}
The final layer of the feature extractor flattens the sparse spiking activation maps $s(t)$ and projects them with $W_D$ to the feature dimension $D$.
Another layer of Parametric LIF dynamics produces the \textbf{feature spike streams} $\mathbf{s}_D(t)$.

\textbf{Pre-training:} During this phase, the feature spike streams are accumulated to feature rates $r_D$ per action sample, which are fed into a linear classifier.
We train this network on 38 action classes (76\% of the THU\textsuperscript{E-ACT}-50 dataset) using cross-entropy loss on a GPU via backpropagation through time and surrogate gradients~\cite{neftci2019}.
During pretraining, we apply common augmentations to the event streams, including random spatial shifts, horizontal flips, and zooms, as well as dropout on the spiking activations $s(t)$.
The pretrained feature extractor is frozen during the subsequent continual learning phase. 

\textbf{Loihi 2 mapping:} We fuse the batch normalization parameters to $Conv_{3\times3}$ weights and neuron biases $b(t)$, and partition and map it to Loihi 2 using Intel's Lava SW framework\footnote{\hyperlink{Lava: A Software Framework for Neuromorphic Computing}{https://github.com/lava-nc/lava}: here the proprietary pieces of this library accessible by INRC community are used}.
The feature extractor allocated 76 of 128 neurocores, leaving the remainder for the aggregation, normalization, and CLP-Loihi modules.

\subsection{CLP-Loihi Module}
The original CLP algorithm~\cite{hajizada2024} and its SNN extension CLP-SNN~\cite{hajizada2025} were designed for static spatial inputs (e.g., images).
Action recognition, therefore, requires integrating information over the duration of the action into a feature vector.
Additionally, CLP-SNN expects input vectors to be L2-normalized. To address these requirements, we have extended CLP-SNN on Loihi 2, resulting in a new version called CLP-Loihi.

\textbf{Temporal Aggregation Layer:}  We introduce this layer to integrate information over time.
The Spiking CNN outputs a binary feature spike stream $s_D(t)$ over the duration of the action clip $T$, where each time step $t$ corresponds to one time bin of the input. 
We feed these into a layer of non-leaky Integrate-and-Fire neurons that accumulate membrane potential over $T$:
\[
V_D(T) = \sum_{t=0}^{T} s_D(t)
\]
At the end of the clip ($t=T$), these neurons discharge their accumulated value as a vector of ``graded spikes''.
Graded spikes can contain up to 24 bits of activation values, and thus act more akin to conventional vectors used in linear algebra, as opposed to binary-valued spikes.
This produces a single action-level feature descriptor $\mathbf{x}_{\text{action}} \in \mathbb{R}^D$ that encapsulates the entire spatiotemporal sequence.

\textbf{Normalization Layer:} Feature vectors extracted by the temporal aggregation layer need to first be normalized to be incorporated into the CLP algorithm. 
Loihi 2 has no native division instruction, so we implement L2-normalization via a fixed-point fast inverse square root algorithm using Loihi 2's microcode engine~\cite{morris2020computing}. The implementation uses two neuron populations: a primary normalization layer (NL) and a single inverse-square-root neuron (ISRN). The NL neurons receive the input vector at their primary input channels via a 1-to-1 connection; each neuron squares the corresponding value and sends the result to the ISRN via its secondary output channel. The ISRN thus receives the sum of squares, calculates the inverse square root via a fixed-point lookup table, and broadcasts the result to all the NL neurons' secondary input channels. Upon receiving the inverse square root value, each NL neuron multiplies each element of the vector by the inverse square root. Finally, the normalized vector is broadcast downstream from the NL's primary output channel to the CLP-SNN.

\textbf{Continual Learning}: CLP-SNN's self-normalizing three-factor rule is designed for Loihi 2's on-chip learning engine~\cite{hajizada2025}. 
Weight updates require only local pre-synaptic spike, post-synaptic spike, and reward signal, which are all available on-chip near the compute, avoiding long-distance memory read-writes. 
Only the weights of the active/winner prototype neuron are updated. 
The self-normalization term of the learning rule eliminates the need for explicit weight normalization, which would require reading the neuron's weight vector into Loihi's embedded CPUs, normalizing it, and rewriting it. 
Instead, each synapse regularizes itself based on local information thanks to a self-normalizing rule~\cite{hajizada2025}.
The normalized aggregated vector $\mathbf{x}_{\text{action}}$ serves as the input to the CLP plasticity layer, enabling the algorithm to learn prototype vectors representing \emph{whole actions}.
While the original CLP algorithm can update previous prototypes, CLP-Loihi creates a new prototype when an error or a novel action occurs, but they are not subsequently adapted.
This simplified mapping enables us to leverage Loihi 2's learning acceleration for CLP's single-shot learning without modifying the core plasticity rule.

\begin{table}[t]
\caption{\textbf{Cross-platform iso-algorithm comparison} using \textbf{CLP-Loihi} as learning algorithm with frozen feature extractors. Final accuracy is reported for 10-shots, while efficiency measurements are for the entire architecture when processing one action clip.}
\label{table:benchmark}
\centering
\begin{small}
\begin{sc}
\begin{tabular}{lcccccr}
\toprule

Platform       & Feature Extractor & Online                 & \makecell{Final Incr.\\ Acc (\%)} & \makecell{Latency\\(\unit{\milli\second})} & \makecell{Energy \\(\unit{\milli\joule})}&
\makecell{EDP\\(\unit{\milli\joule\second})}\\
\midrule
 Orin Nano   & 2D CNN + GRU   & $\surd$       & 73.0$\pm$3.6& 80   & 427 & 34.2 \\
 Orin Nano   & 3D CNN    &  $\times$          & \textbf{74.9$\pm$1.3} & 9.0    & \textbf{2.0} & \textbf{0.018}\\
 Loihi 2       & Spiking 2D CNN & $\surd$ & 70.4$\pm$2.0& \textbf{4.79} & 3.78 & \textbf{0.018}\\
\bottomrule
\end{tabular}
\end{sc}
\end{small}
\end{table}

\section{Experimental Setup}
\label{sec:exp_setup}
\textbf{Dataset and Continual Learning Protocol:}
THU\textsuperscript{E-ACT}-50 dataset has 50 action classes, 10,500 video samples, and real-world conditions, such as lighting variation, viewpoint changes, and occlusions.
Our learning protocol has two phases and hence we split the dataset into 38 base classes and 12 hold-out classes.
The samples from every fourth class are selected for the hold-out classes. 
In the initial offline pretraining phase, we train the spiking CNN on the 38 base classes in a supervised manner for 50 epochs.
We remove the output classification layer and use the spiking CNN's penultimate layer as the feature layer.
In the continual learning phase, we connect the CLP-Loihi module (including temporal aggregation and normalization layers) to this feature extractor and perform online class-incremental learning of the 12 unseen, hold-out action classes.
Online learning proceeds with only a single pass over the data, and no replay buffer is provided.
We test on all classes learned so far after each new class.

\textbf{Cross-Platform Comparison Methodology:}
For benchmarking performance on neuromorphic and conventional hardware, we select the Nvidia Jetson Orin Nano as the comparison platform. Following the same pretraining procedure, we train a 2D CNN plus a GRU layer and a 3D CNN with ReLU activations for each input frequency. All variants have the same number of layers and feature dimensions as the spiking CNN. The action-level feature descriptors $x_\text{action}$ are determined by averaging frame-level features over the sample duration. For comparison of continual learning performance we select NCM~\cite{mensink2013dncm}, Replay~\cite{hayes2019}, SLDA~\cite{hayes2020} and CLP~\cite{hajizada2024} as learning algorithms. We implement the CNN feature extractors and learning algorithms on Jetson Orin Nano. Furthermore, we set up a sequence of cross-platform comparison experiments in which one algorithmic pipeline stage remains fixed to create meaningful insights. 

\begin{figure}[t]
\includegraphics[width=\textwidth]{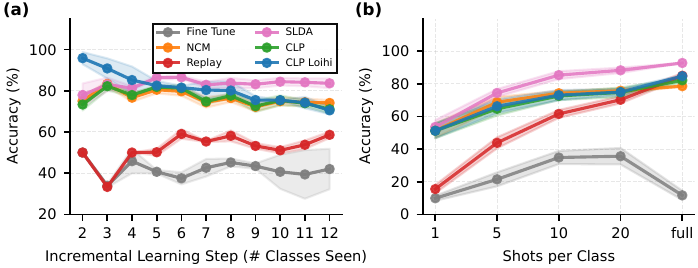}
\caption{Continual learning accuracy on THU\textsuperscript{E-ACT}-50 (12 hold-out classes). (a) Cumulative accuracy after each class-incremental step (10-shot). (b) Final accuracy versus shots per class. CLP-Loihi runs on Loihi 2; all others on Jetson Orin Nano.}
\vspace{-5pt}
\label{fig:acc_trend}
\end{figure}

\section{Results}
\label{sec:results}

\begin{figure}[t]
\centerline{\includegraphics[width=0.9\textwidth]{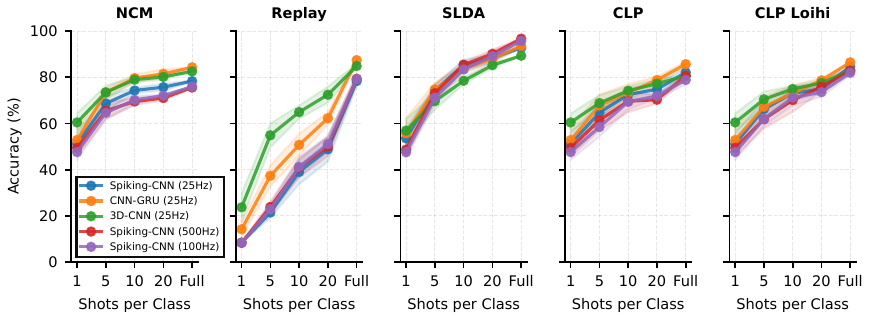}}
\caption{Comparison of different feature extraction and learning method combinations. }
\vspace{-10pt}
\label{fig:scnn_classifiers}
\end{figure}

\textbf{End-to-End Continual Learning Performance:}
After pretraining the spiking CNN and baseline feature extractors on the 38 basic classes, all models reach between 85-90\% test accuracy on these classes.
In \cref{fig:acc_trend}(a), we show the evolution of accuracies over learned classes for (10-shot) class-incremental learning of the 12 hold-out classes and demonstrate that naive fine-tuning catastrophically forgets as new classes are learned, while CLP and other CL methods retain knowledge of previous classes to various degrees.
In \cref{fig:acc_trend}(b), we report the final accuracies for the same experiment, but for different numbers of shots (class presentations).
CLP-Loihi shows competitive performance on par with CLP and NCM, outperforming Fine Tune and Replay (except for the full set), while slightly being outperformed by SLDA ($\sim$5-10\%). While these used the 25Hz spiking CNN, \cref{fig:scnn_classifiers} evaluates the results also for the other feature extractors.
For CLP-Loihi and CLP the spiking CNNs lead to similar accuracies as the baseline CNNs, only marginally lower for small number of shots per class. For SLDA the SNNs perform on par or even outperform the 3D CNN variant.

\textbf{Iso-Algorithm Cross-Platform Comparison:}
To rigorously assess practical deployment trade-offs, we compare neuromorphic and conventional platforms at iso-algorithm operating points with the same number of deep CNN layers and learning algorithm, eliminating confounds from differences in model capacity and learning. The key result is latency and energy reduction at the iso-algorithm. In \cref{table:benchmark} we present these full pipeline results for the 2D CNN + GRU and 3D CNN combined with CLP-Loihi and benchmarked on Jetson Orin Nano and our CLANE on Loihi 2. While the final incremental accuracies for 10 shots per class are almost equal, our pipeline is \textbf{16× faster} and \textbf{$>$100× more efficient} than the 2D CNN + GRU pipeline on Orin Nano. Compared to the 3D CNN our pipeline achieves a speed-up of 2×, while consuming 1.9× more energy, yielding the same energy-delay product. This result shows that for batched processing of sequences like for the 3D CNN the GPU is very efficient. In a real-time, streaming deployment scenario the 3D CNN pipeline comes with additional latency or computational effort for sliding window processing.

\textbf{Learning Algorithm Landscape:}
In \cref{table:benchmark2}, we present benchmarking results for multiple CL algorithms applied to features from the same frozen spiking CNN, isolating the contribution of the learning module.
SLDA achieves the highest accuracy but requires maintaining a shared $D\times D$ covariance matrix, updated via a per-sample outer product and inverting it via SVD -- operations that require globally shared dense memory and sequential computation, with no local-synapse equivalent, making SLDA fundamentally incompatible with Loihi 2's architecture.
Replay requires storing and randomly sampling exemplars from a replay buffer, necessitating large off-chip DRAM accesses (850 MB in our setup) that would eliminate energy-efficiency gains from on-chip learning.
NCM is a simple and fast method on the GPU, but requires Euclidean distance computation across all stored prototypes, which is not compatible with the synaptic computation primitives of the Loihi 2.
Of the algorithms evaluated, CLP-SNN is the only one whose inference and local three-factor rule satisfy Loihi 2's architectural constraints. 
At this iso-algorithm operation point for CLP-Loihi, Loihi 2 provides \textbf{95× energy reduction}, and \textbf{10× learning speedup} for the learning module, at a 2.6\% accuracy cost from neuromorphic deployment, attributable to 8-bit integer precision and the simplified single-shot prototype creation policy.

\begin{table}[t]
  \begin{minipage}[t]{0.48\textwidth}
    \caption{\textbf{Algorithmic comparison}. Final 10-shot accuracy, latency, and energy for the learning algorithms on Jetson Orin Nano. Deployment on Loihi 2 is specified with a dagger (${\dag}$).}
    \label{table:benchmark2}
    \centering
    \begin{tabular}{lccr}
\toprule

\makecell{Learning\\algorithm}  & \makecell{Final\\Acc (\%)} & \makecell{Latency\\(\unit{\milli\second})} &  \makecell{Energy\\(\unit{\milli\joule})} \\
\midrule
NCM       & $74.3 \pm 1.2$&             0.8  & 3.8  \\
Replay    & $61.5 \pm 2.4$&             3.3  & 15.8  \\
SLDA      & $\mathbf{85.3 \pm 2.4}$&    14.1 & 111.1  \\
CLP       & $\underline{72.5 \pm 2.5}$& 3.7  & 17.7 \\
CLP-Loihi & $73.0 \pm 3.1$&             2.4  & 11.4 \\
CLP-Loihi$^{\dag}$   & 70.4$\pm$ 2.0&        \bf{0.24}   & \bf{0.12} \\
   
\bottomrule
\end{tabular}
  \end{minipage}
  \hfill
  \begin{minipage}[t]{0.48\textwidth}
    \caption{\textbf{Benchmarking of feature extractors} Spiking CNN (Loihi 2) vs. CNN + GRU (Orin) for different input frequencies. We report final 10-shot accuracy, latency and energy.}
    \label{table:benchmark3}
    \centering
    \begin{tabular}{lcccr}
\toprule

\makecell{Freq \\ (\unit{\hertz})} & HW & \makecell{Final\\Acc (\%)} & \makecell{Latency\\(\unit{\milli\second})} &  \makecell{Energy\\(\unit{\milli\joule})}\\
\midrule
25   & Loihi 2& 70.4$\pm$2.0& \textbf{1.87} & \textbf{1.5} \\
25   & Orin & 73.0$\pm$3.6& 64   & 341 \\
100  & Loihi 2& 68.5$\pm$1.4& 7.5  & 5   \\
100  & Orin & 72.2$\pm$2.1& 208  & 1229  \\
500  & Loihi 2& 69.6$\pm$1.8& 37.3 & 25  \\
500  & Orin & \textbf{73.3$\pm$1.1}& 970  & 5629 \\
   
\bottomrule
\end{tabular}
  \end{minipage}
  \vspace{-10pt}
\end{table}

\textbf{Feature Extractor Analysis:}
Human actions evolve on timescales of 100 ms–2 s, making 40 ms resolution sufficient for capturing motion dynamics~\cite {gao2023}. Higher temporal rates (e.g., 500 Hz used in~\cite{neumeier2025}) provide marginal accuracy gains ($<2\%$) for action recognition but substantially increase computational cost on conventional hardware. The results shown in \cref{table:benchmark3} validate this observation. The lower the input frequency, the lower the latency and energy consumption. Both metrics almost scale linearly with input frequency. When comparing across platforms, the spiking CNNs on Loihi 2 provide \textbf{$\sim$30× latency} and \textbf{$\sim$230× energy reduction} compared to the 2D CNN + GRU on Jetson Orin Nano.

\section{Discussion and Conclusion}
\label{sec:discussion}
Our results demonstrate that architectural fit between event cameras, SNNs, local learning, and neuromorphic hardware creates synergistic efficiency gains.
GPU's SIMD execution model demands dense tensor operations regardless of input sparsity, backpropagation requires storing all intermediate activations regardless of update locality, and von Neumann architectures incur DRAM access regardless of computational sparsity.
CLANE eliminates all three bottlenecks simultaneously: event sparsity propagates from the sensor through SNN computation to on-chip weight updates.
The neuromorphic stack dominates in the sparse, online, power-constrained regime -- precisely the deployment scenario for edge continual learning from event cameras.
For dense data, offline training, or unconstrained power budgets, GPU advantages in batch processing and model capacity may outweigh neuromorphic efficiency and understanding this regime boundary is critical for system designers.

Several limitations bound the current system. The spiking CNN feature extractor is frozen after pretraining, which may limit generalization to actions with visual statistics far from the base training classes.
CLP-Loihi creates new prototypes on error events but does not subsequently adapt existing ones, which could cause prototype accumulation in very long deployment sequences.

To conclude, we presented CLANE, an end-to-end continual learning system for event cameras deployed on Loihi 2, integrating a 2D spiking CNN with CLP-SNN for online class-incremental action recognition.
Two novel Loihi 2 modules -- temporal aggregation \& fixed-point normalization layers -- extend CLP-SNN to spatiotemporal tasks for the first time.
Evaluated on THU\textsuperscript{E-ACT}-50 under real-world conditions, CLANE achieves competitive 10-shot incremental accuracy (70.4\%) while delivering substantial efficiency gains ($100\times$ energy reduction and $16\times$ speed-up) compared to a same-sized sequential CNN+GRU pipeline on edge GPU.
This work establishes neuromorphic hardware as a practical substrate for always-on, continually adapting perception systems and provides a benchmarking methodology that the neuromorphic CL field currently lacks.



%
%
%
\bibliographystyle{splncs04}
\bibliography{main}

@article{stewart2020online,
  title={Online few-shot gesture learning on a neuromorphic processor},
  author={Stewart, Kenneth and Orchard, Garrick and Shrestha, Sumit Bam and Neftci, Emre},
  journal={IEEE JETCAS},
  volume={10},
  number={4},
  pages={512--521},
  year={2020},
  publisher={IEEE}
}

@article{vicente2025,
  title={Spiking Neural Networks for event-based action recognition: A new task to understand their advantage},
  author={Vicente-Sola, Alex and others},
  journal={Neurocomputing},
  volume={611},
  pages={128657},
  year={2025},
  publisher={Elsevier}
}

@article{gao2023,
  title={Action recognition and benchmark using event cameras},
  author={Gao, Yue and others},
  journal={IEEE TPAMI},
  volume={45},
  number={12},
  pages={14081--14097},
  year={2023},
  publisher={IEEE}
}

@inproceedings{neumeier2025,
  title={{EEvAct: Early Event-Based Action Recognition with High-Rate Two-Stream Spiking Neural Networks}},
  author={Neumeier, Michael and others},
  booktitle={ICONS},
  pages={41--48},
  year={2025}
}

@article{blegiers2023,
  title={{EventTransAct: A Video Transformer-Based Framework for Event-Camera Based Action Recognition}},
  author={Tristan de Blegiers and Ishan Rajendrakumar Dave and Adeel Yousaf and Mubarak Shah},
  journal={IROS},
  year={2023},
  pages={1-7},
}

@article{innocenti2020,
  title={{Temporal Binary Representation for Event-Based Action Recognition}},
  author={Simone Undri Innocenti and Federico Becattini and Federico Pernici and A. Bimbo},
  journal={ICPR},
  year={2020},
  pages={10426-10432},
}

@inproceedings{amir2017dvsGestures,
  title={A low power, fully event-based gesture recognition system},
  author={Amir, Arnon and others},
  booktitle={Proceedings of the IEEE CVPR},
  pages={7243--7252},
  year={2017}
}

@article{hajizada2025,
  title={{Real-time Continual Learning on Intel Loihi 2}},
  author={Hajizada, Elvin and others},
  journal={arXiv preprint arXiv:2511.01553},
  year={2025}
}

@inproceedings{hajizada2024,
  title={{Continual Learning for Autonomous Robots: A Prototype-based Approach}},
  author={Hajizada, Elvin and Swaminathan, Balachandran and Sandamirskaya, Yulia},
  booktitle={IROS},
  year={2024},
  organization={IEEE}
}

@article{minhas2025continual,
  title={Continual learning with neuromorphic computing: Foundations, methods, and emerging applications},
  author={Minhas, Mishal Fatima and Putra, Rachmad Vidya Wicaksana and Awwad, Falah and Hasan, Osman and Shafique, Muhammad},
  journal={IEEE Access},
  year={2025},
  publisher={IEEE}
}

@article{davies2018loihiLearning,
   author = {Mike Davies and others},
   doi = {10.1109/MM.2018.112130359},
   issn = {02721732},
   issue = {1},
   journal = {IEEE Micro},
   pages = {82-99},
   title = {{Loihi: A Neuromorphic Manycore Processor with On-Chip Learning}},
   volume = {38},
   year = {2018},
}

@inproceedings{orchard2021loihi,
  title={Efficient neuromorphic signal processing with loihi 2},
  author={Orchard, Garrick and others},
  booktitle={2021 IEEE Workshop on SiPS},
  pages={254--259},
  year={2021},
  organization={IEEE}
}

@article{schuman2022neuromorphicSurvey,
  title={Opportunities for neuromorphic computing algorithms and applications},
  author={Schuman, Catherine D and others},
  journal={Nature Computational Science},
  volume={2},
  number={1},
  pages={10--19},
  year={2022},
  publisher={Nature Publishing Group US New York}
}

@article{yamazaki2022snn,
  title={Spiking neural networks and their applications: A review},
  author={Yamazaki, Kashu and Vo-Ho, Viet-Khoa and Bulsara, Darshan and Le, Ngan},
  journal={Brain sciences},
  volume={12},
  number={7},
  pages={863},
  year={2022},
  publisher={MDPI}
}

@article{fang2021,
  title={{Incorporating Learnable Membrane Time Constant to Enhance Learning of Spiking Neural Networks}},
  author={Wei Fang and others},
  journal={ICCV},
  year={2021},
  pages={2641-2651},
}

@article{neftci2019,
  author={Neftci, Emre O. and Mostafa, Hesham and Zenke, Friedemann},
  journal={IEEE Signal Processing Magazine}, 
  title={{Surrogate Gradient Learning in Spiking Neural Networks: Bringing the Power of Gradient-Based Optimization to Spiking Neural Networks}}, 
  year={2019},
  volume={36},
  number={6},
  pages={51-63},}

@article{panda2017asp,
  title={{Asp: Learning to forget with adaptive synaptic plasticity in spiking neural networks}},
  author={Panda, Priyadarshini and Allred, Jason M and Ramanathan, Shriram and Roy, Kaushik},
  journal={IEEE JETCAS},
  volume={8},
  number={1},
  year={2017},
  publisher={IEEE}
}

@inproceedings{pes2024activedend,
  title={Active dendrites enable efficient continual learning in time-to-first-spike neural networks},
  author={Pes, Lorenzo and Luiken, Rick and Corradi, Federico and Frenkel, Charlotte},
  booktitle={AICAS},
  year={2024},
  organization={IEEE}
}

@article{han2023DSD-SNN,
  title={Enhancing efficient continual learning with dynamic structure development of spiking neural networks},
  author={Han, Bing and Zhao, Feifei and Zeng, Yi and Pan, Wenxuan and Shen, Guobin},
  journal={arXiv preprint arXiv:2308.04749},
  year={2023}
}

@inproceedings{proietti2023snnreplay,
  title={Memory replay for continual learning with spiking neural networks},
  author={Proietti, Michela and Ragno, Alessio and Capobianco, Roberto},
  booktitle={2023 IEEE 33rd International Workshop on MLSP},
  pages={1--6},
  year={2023},
  organization={IEEE}
}

@article{minhas2025replay4ncl,
  title={{Replay4NCL: An Efficient Memory Replay-based Methodology for Neuromorphic Continual Learning in Embedded AI Systems}},
  author={Minhas, Mishal Fatima and others},
  journal={arXiv preprint arXiv:2503.17061},
  year={2025}
}

@article{mensink2013dncm,
  title={Distance-based image classification: Generalizing to new classes at near-zero cost},
  author={Mensink, Thomas and Verbeek, Jakob and Perronnin, Florent and Csurka, Gabriela},
  journal={IEEE transactions on pattern analysis and machine intelligence},
  volume={35},
  number={11},
  pages={2624--2637},
  year={2013},
  publisher={IEEE}
}

@article{hayes2019,
  title={{REMIND Your Neural Network to Prevent Catastrophic Forgetting}},
  author={Tyler L. Hayes and Kushal Kafle and Robik Shrestha and Manoj Acharya and Christopher Kanan},
  journal={ECCV},
  year={2019}
}

@InProceedings{hayes2020,
    author = {Hayes, Tyler L. and Kanan, Christopher},
    title = {{Lifelong Machine Learning With Deep Streaming Linear Discriminant Analysis}},
    booktitle = {CVPR Workshop},
    month = {June},
    year = {2020}
}

@article{gallego2020eventSurvey,
  title={Event-based vision: A survey},
  author={Gallego, Guillermo and Delbr{\"u}ck, Tobi and Orchard, Garrick and Bartolozzi, Chiara and others},
  journal={IEEE TPAMI},
  volume={44},
  number={1},
  pages={154--180},
  year={2020},
  publisher={IEEE}
}

@inproceedings{shen2024towards,
  title={Towards open-world gesture recognition},
  author={Shen, Junxiao and De Lange, Matthias and Xu, Xuhai and others},
  booktitle={2024 IEEE ISMAR},
  pages={1236--1245},
  year={2024},
  organization={IEEE}
}

@article{de2021CLreview,
  title={A continual learning survey: Defying forgetting in classification tasks},
  author={De Lange, Matthias and Aljundi, Rahaf and Masana, Marc and Parisot, Sarah and others},
  journal={IEEE TPAMI},
  volume={44},
  number={7},
  year={2021},
  publisher={IEEE}
}

@article{hayes2022online,
  title={Online continual learning for embedded devices},
  author={Hayes, Tyler L and Kanan, Christopher},
  journal={arXiv preprint arXiv:2203.10681},
  year={2022}
}

@article{french1999catastrophic,
  title={Catastrophic forgetting in connectionist networks},
  author={French, Robert M},
  journal={Trends in cognitive sciences},
  volume={3},
  number={4},
  pages={128--135},
  year={1999},
  publisher={Elsevier}
}

@article{morris2020computing,
    title={Computing Fixed-Point Square Roots and Their Reciprocals Using Goldschmidt Algorithm},
    author={Michael Morris},
    journal={FPGA Related},
    year={2020},
    url={https://www.fpgarelated.com/showarticle/1347.php}
}

@article{fremaux2016neuromodulated,
  title={Neuromodulated spike-timing-dependent plasticity, and theory of three-factor learning rules},
  author={Fr{\'e}maux, Nicolas and Gerstner, Wulfram},
  journal={Frontiers in neural circuits},
  volume={9},
  pages={85},
  year={2016},
  publisher={Frontiers Media SA}
}

@article{lillicrap2020backpropagation,
  title={Backpropagation and the brain},
  author={Lillicrap, Timothy P and Santoro, Adam and Marris, Luke and Akerman, Colin J and Hinton, Geoffrey},
  journal={Nature Reviews Neuroscience},
  volume={21},
  number={6},
  pages={335--346},
  year={2020},
  publisher={Nature Publishing Group UK London}
}

\end{document}